\pdfoutput=1

\documentclass[11pt]{article}

\usepackage[final]{acl}

\usepackage{times}
\usepackage{latexsym}

\usepackage[T1]{fontenc}

\usepackage[utf8]{inputenc}
\usepackage{amsmath}
\usepackage{booktabs}
\usepackage{multirow}
\usepackage{microtype}

\usepackage{inconsolata}

\usepackage{graphicx}
\usepackage{titletoc}
\usepackage{hyperref}
\usepackage{xcolor}
\definecolor{RoyalBlue}{rgb}{0.0, 0.14, 0.4}
\hypersetup{
    colorlinks=true,    
    linkcolor=RoyalBlue,     
    citecolor=RoyalBlue,      
    urlcolor=RoyalBlue       
    }

\usepackage{colortbl}
\usepackage{booktabs}
\usepackage{subcaption}
\usepackage[normalem]{ulem}
\usepackage{amssymb}
\usepackage{enumitem}
\title{Judging the Judges: A Systematic Study of Position Bias in LLM-as-a-Judge}

\author{Lin Shi, Chiyu Ma, Wenhua Liang, Xingjian Diao, Weicheng Ma, Soroush Vosoughi \\
Dartmouth College \\
\small {lin.shi.26@dartmouth.edu}
}

\begin{document}
\maketitle
\begin{abstract}

LLM-as-a-Judge has emerged as a promising alternative to human evaluators across various tasks, yet inherent biases—particularly position bias, the tendency to favor solutions based on their position within the prompt—compromise its reliability. This exploratory study evaluates position bias in LLM judges across pairwise and list-wise comparison settings, introducing three metrics: repetition stability, position consistency, and preference fairness. Our experiments, involving 15 LLM judges across MTBench and DevBench with 22 tasks and approximately 40 solution-generating models, result in over 150,000 evaluation instances. We identify Judge-Level, Candidate-Level, and Task-Level factors contributing to bias. The findings confirm that position bias is not due to random chance and varies significantly across judges and tasks. While position bias is weakly influenced by the length of prompt components, it is strongly affected by the quality gap between solutions. Our agreement and disagreement analysis among judges further provides insights into the distribution of judging difficulty across the dataset, and highlights the potential for dataset modifications.  

\end{abstract}

\section{Introduction}


\begin{figure*}[tb]
\begin{center}
\includegraphics[width=1\linewidth]{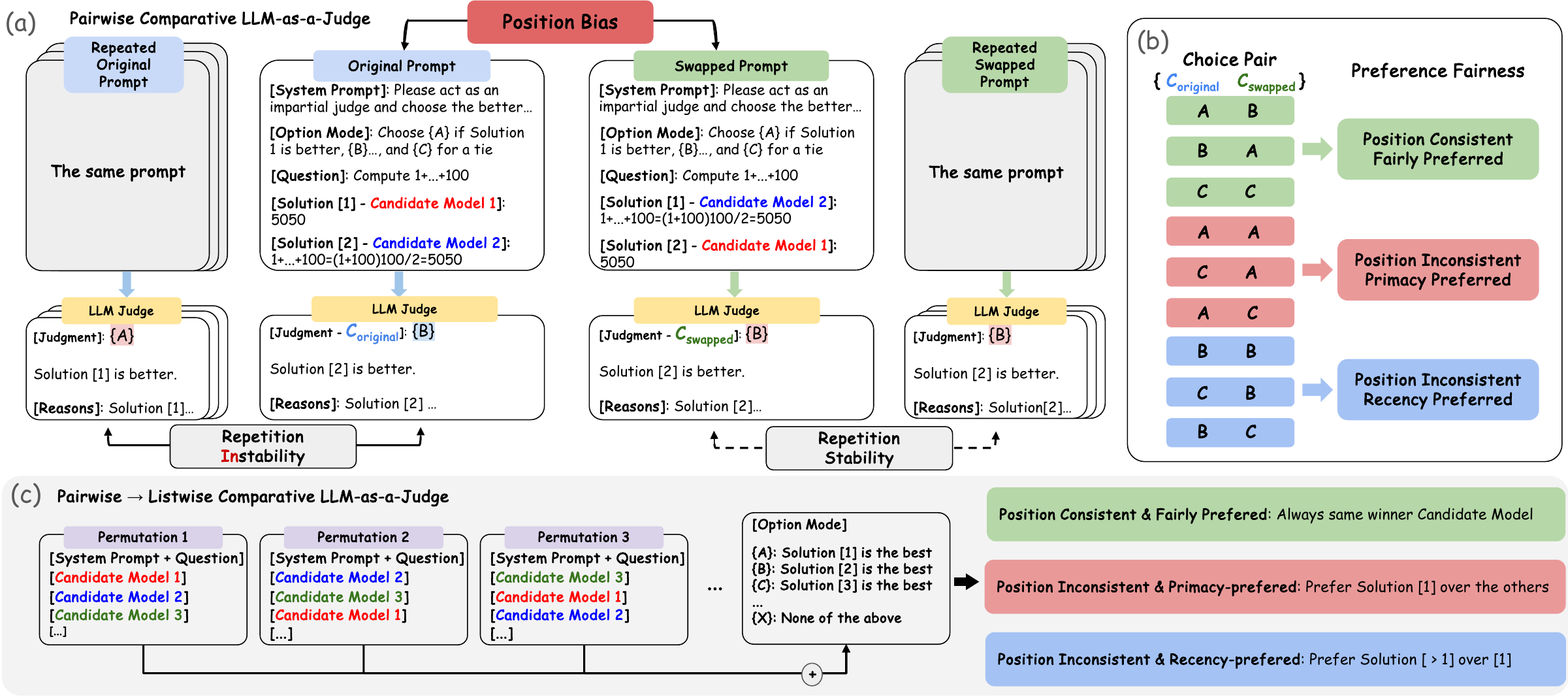}
\end{center}
\vspace{-0.45cm}
\caption{Overview of our experiment settings: (a) Position bias is observed when LLM judges consistently favor a specific position rather than evaluating the content, with repeated trials ensuring the deviations are not due to random variations. (b) Preference fairness is defined and measured through the distribution of choice pairs to assess the fairness of judgments. (c) The settings are extended from pairwise comparisons to list-wise comparisons, involving evaluations of more than two candidate models.}
\vspace{-0.7cm}
\label{demo}
\end{figure*}


In recent years, Large Language Models (LLMs) have emerged as evolutionary technologies, gathering global interest and stimulating substantial research into their applications. Evaluating LLMs has received increasing attention due to their advancing capabilities across diverse fields. While human assessment is considered the gold standard for aligning with human preferences, it lacks scalability in extensive evaluations \cite{zeng2023evaluating, karpinska2021perils}. To automate evaluations and reduce reliance on costly human evaluators, the LLM-as-a-Judge methodology emerged as a promising alternative across various tasks. Despite a high level of agreement with human judgments \cite{zheng2024judging,li2024devbench,zhu2023judgelm}, inherent biases, especially \textbf{position bias}, have undermined the accuracy, fairness, and reliability of these LLM evaluators. 

\textbf{Position bias} refers to the tendency of LLM judges to favor certain positions within prompt components rather than evaluating content objectively, as shown in Fig.~\ref{demo} (a). This bias has been observed across various types of LLM judges \cite{qin2024large, li2023split}, raising concerns about their reliability. Prior studies have identified position bias alongside other biases and assessed its impact \cite{zheng2024large,zheng2024judging,zeng2023evaluating}, but these investigations remain largely preliminary and lack a focused, systematic exploration. Although mitigation strategies have been proposed, they often suffer from incomplete bias removal \cite{guo2024bias}, added complexity \cite{li2024split,khan2024debating,chua2024biasaugmented}, the introduction of new biases \cite{ohi2024likelihood}, inconsistent effectiveness \cite{gallegos2024bias}, or impracticality for closed-source models requiring access to model internals \cite{wang2025eliminating,yu2025mitigate}. Moreover, most empirical studies primarily examine position bias in pairwise settings \cite{wang2023large,zheng2024judging}, leaving its underlying factors and behavior in more complex list-wise paradigms underexplored. These gaps underscore the need for a more systematic, in-depth exploratory analysis of position bias to better understand its origins, manifestations, and implications.

In this study, we provide an in-depth and systematic investigation into position bias within the context of LLM-as-a-Judge. While evaluating with \textbf{Position Consistency} \cite{zheng2024judging}, we further introduce two novel metrics: \textbf{Preference Fairness} and \textbf{Repetition Stability}. Specifically, we move beyond simply assessing Position Consistency by incorporating Preference Fairness, which provides deeper insights into the specific answer directions where models exhibit unfair preferences. Additionally, the measurement of Repetition Stability ensures that the observed position bias in the given model and tasks is not due to random variations, thus strengthening the reliability of the findings.

To investigate the underlying factors contributing to position bias, we categorized these factors into three levels: \textbf{Judge-Level}, \textbf{Candidate-Level}, and \textbf{Task-Level}. Our experiments are primarily conducted on pairwise comparisons, as LLM judges demonstrate superior performance in this setting. We further extend our study to more complicated list-wise comparison settings, involving evaluations of more than two candidate models by LLM judges. \textbf{Our findings reveal several key insights:} 1. The position bias of capable LLM judges is not a result of random variations. 2. There is a high volatility in the direction of preference, even within the same LLM judge when applied to different tasks. 3. Differences in answer quality among candidate models significantly influence position consistency. 4. Position bias is very weakly correlated with the length of prompts generated by the candidate models.

Building on these findings, we conduct an agreement analysis among the LLM judges. The results reveal that, although measures of position consistency may appear similar in general, judgments on specific instances vary significantly among LLM judges, even when they demonstrate comparable capabilities. \textbf{Instances where numerous LLMs agree are generally easier to judge, whereas instances with disagreements are more challenging to evaluate and more prone to position bias}. This analysis provides insights into the distribution of judging difficulty across the dataset and highlights the potential for dataset modifications by incorporating more instances that are either easier or more difficult to judge. Future work could explore how to measure the likelihood of position bias arise from the datasets by identifying and quantifying such hard-to-judge instances before implementing LLM judges.

\section{Evaluation Settings \& Definitions}

We begin by outlining the settings for pairwise and list-wise comparisons employed in our experiments for LLM-as-a-Judge. Following this, we define the three metrics used in our evaluation: Position Consistency (PC), Preference Fairness (PF), and Repetition Stability (RS). Finally, we provide a detailed description of the factors we found that are related to position bias at the Judge-Level, Candidate-Level, and Task-Level. Our exploratory study was conducted post hoc, meaning the LLM judgments were collected first, and the factors influencing position bias were then identified and analyzed.

\subsection{Pairwise \& List-wise Comparison}

\textbf{Pairwise Comparison:} In the context of pairwise comparison, LLM judges are tasked with selecting the better solution provided by two candidate models in response to a given task question. As shown in Fig. \ref{demo} (a), the system prompt, option choices, task question, and solutions from two candidate models (\textit{original prompt}) are presented to the LLM judges to select the better solution. The experiment is conducted in a double-blind setting. The identities of the candidate models are hidden from the LLM judges, and the candidate models are unaware that their solutions will be compared to another model when answering the question. Then, the prompt with solutions in a swapped position (\textit{swapped prompt}) is given to the same LLM judge again, which results in a judgment pair. If the LLM judge consistently favors the same solution regardless of the swapped position, it is considered position consistent. Conversely, if the LLM judge selects different winners, position bias is observed, with the preference direction being either primacy (e.g. always choose $\{A\}$) or recency (e.g.  always choose $\{B\}$). Example of measuring preference fairness with specific choice pairs is shown in Fig. \ref{demo} (b). To accommodate the possibility of ties, various option modes are employed: \textbf{Two-Option mode} restricts LLM judges to choosing between two options, labeled {A} for the first candidate and {B} for the second. \textbf{Three-Option mode} introduces an additional choice, {C}, allowing LLM judges to indicate a tie if neither solution is preferable, as illustrated in Fig. \ref{demo} (a). These option modes were explicitly specified in the system prompts to ensure clear guidance for the decision-making process of the LLM judges.

\textbf{List-wise Comparison:} Unlike pairwise settings, where LLM judges select the superior solution from two candidates, list-wise comparative approaches involve evaluating three or more candidates simultaneously, as shown in Fig. \ref{demo} (c). For efficiency, we prompt LLM judges to select the best candidate rather than ranking the entire list. The ``swapped setting" used in pairwise evaluations is generalized to order permutations for list-wise judgments, ensuring that each candidate appears in every possible position exactly once. For a list of $p$ candidates, this results in $p$ permutations. In the $i^{th}$ permutation, the $i^{th}$ candidate is set to appear in the first position. Additionally, an option is provided to account for ties, allowing LLM judges to indicate if there is no certainly superior solution.

\subsection{Evaluation Metrics}
In our study, we first verify whether capable LLM judges exhibit high repetition stability and then evaluate their position bias in terms of position consistency and preference fairness. The metrics are introduced as follows:

\textbf{Repetition Stability ($RS$)} evaluates the reliability of LLM judges when presented with identical queries multiple times. It is essential to determine whether the judgments of LLMs, and consequently the observations of position bias, stem from a consistent evaluation pattern or by random variations. We measure this by calculating the percentage of the most frequent selections across multiple trials for each query, aggregated from all queries within each dataset. This metric is formalized as
\begin{equation}
\begin{aligned}
   RC &= \frac{1}{N} \sum_{j=1}^N \frac{1}{n_j} \max_{k \in S} \left\{ |C_k^j| \right\},
   \end{aligned}
\end{equation}
where $S = \{A, B, C, \dots\}$ refers to the set of choice options depending on the option mode, $|C_{k}^j|$ denotes the counts of each choice option selected by the judge for the $j^{th}$ query, $n_j$ represents the total number of repeating trials for that query, and $N$ is the total number of queries. The value of $RS$ ranges from a small positive value depending on the option mode, indicating completely random decisions, to 1.0, indicating perfect stability.

\textbf{Position Consistency ($PC$)} quantifies how frequently LLM judges prefer the same solution after the order of solutions is permuted. It is calculated as the ratio of consistent evaluation series to the total number of valid evaluations, where a series is deemed consistent if the LLM judge prefers the same winning solution across permutations.
Formally, it is calculated as
\begin{equation}
\begin{aligned}
    PC &= \frac{1}{n} \sum_{j=1}^n \mathbf{1}_{\{(C_{1}^j, \dots, C_{p}^j, ) \in V\}},
\end{aligned}
\end{equation}
where $V$ is the set of choices that correspond to position consistency, and $(C_{1}^j, \dots, C_{p}^j)$ denotes the judgment series for the $j^{th}$ query when there are $p$ candidate solutions in the list, and $n$ represents the number of prompt series. An example of such series of choices under pairwise comparison setting can be found in Fig. \ref{demo} (b). This formula aims to provide a direct measure of a LLM judge's position bias and has been widely used in previous studies for its simplicity. 

\textbf{{Preference Fairness ($PF$)}} measures the extent to which LLM judges favor certain solution positions. In pairwise comparisons, an LLM judge may exhibit a preference for either primacy or recency. These terms replace the more verbose ``\textit{preference/bias for the first/second candidate model}" used in previous studies \cite{zheng2024judging}, ensuring clarity and generalization for future research. The examples of such preferences are demonstrated in Fig. \ref{demo} (b). Previous studies proposed two common ways to measure the preference fairness. One way is to count the primacy-preferred and recency-preferred judgment pairs, which we termed as primacy-count-number ($pcn$) and recency-count-number ($rcn$). The counts are then normalized by the total number of prompt pairs \citep{zheng2024judging,zhu2023judgelm}. However, \textbf{the sensitivity of this measurement highly depends on the size of dataset}, making comparisons across datasets unreliable, especially when the number of questions and instances varies for each task. 

Alternatively, instead of normalizing over the complete dataset, studies like \citep{li2023prd, liusie2024llm} treat position inconsistent evaluation instances independently. They calculate the percentages of primacy-preferred and recency-preferred judgment pairs relative to the total number of position inconsistent pairs. We denote these as \textit{inconsistent primacy rates ($ipr$)} and \textit{inconsistent recency rates ($irr$)}, where \(ipr+irr=1\). However, this approach overlooks the fact that ``position consistent judgments are also preference fair'', which leads to \textbf{overly penalizing highly consistent LLM-judges}. 

To overcome these limitations, we introduce a more granular and scalable measurement that combines the strengths of both methods, to assess preference fairness. The $PF$ score is formally calculated by 
\begin{equation}
\begin{aligned}
    PF &= \frac{PF_{\text{raw}} - S^-_{\min}}{S^+_{\max} - S^-_{\min}} \times 2 - 1, \\
   PF_{\text{raw}} &= (rcn \times irr) - (pcn \times ipr).
\end{aligned}
\end{equation}
where $S^-_{min}$ and $S^+_{max}$ are the minimum and maximum achievable $PF_{raw}$ scores for each judge on each task, respectively. This min-max scale ensures comparability across datasets by accounting for the range of achievable scores and centering the scale around zero. The $PF$ score is interpreted as follows: 
\[
\scalebox{0.75}{$
PF = \begin{cases}
    1, & \text{if } PC=0 \text{ and entirely recency-preferred} \\
    x \in (0, 1), & \text{Recency-preferred} \\
    0, & \text{Preference Fair} \\
    x \in (-1, 0), & \text{Primacy-preferred} \\
    -1, & \text{if } PC=0 \text{ and entirely primacy-preferred}
\end{cases}
$}
\]

To extend this metric to list-wise comparisons, we employed a `one vs. all' approach, defining primacy preference as favoring the first candidate solution while classifying all others as recency-preferred. This straightforward extension of the $PF$ computation maintains consistency with pairwise setups. By providing a single and comprehensive metric that applies to all evaluation instances and list-wise settings, our proposed $PF$ score ensures sensitivity across datasets, regardless of variations in the number of questions or instances, offering a significant improvement over previous methods.



\subsection{Factors Affecting Position Bias}
\label{sec:factors_affecting_position_bias}

To investigate the factors influencing position bias in LLM judges, we categorized these factors into three groups: \textbf{Judge-level}, \textbf{Candidate-level,} and \textbf{Task-level} factors. Each group includes specific factors, that we hypothesize, may impact position bias, which we explore through a series of experiments. Table \ref{tbl:impact-factors} lists the five factors we analyzed in this study.

Among the influencing factors, we selected ``familial property'' for Judge-level factors, as it reflects similar model sizes or training specifics, which are often proprietary and not publicly accessible for closed-source capable models. The familial categories of the models used in our studies are (1) \emph{GPT}, (2) \emph{Claude}, (3) \emph{Gemini}, and (4) \emph{Llama} allowing for straightforward grouping by company and version. More details and discussions about the familial property can be found in Appendix Sec. \ref{appendix:agreement_analysis}.

\textbf{Answer quality gap:} While prior studies \cite{wang2023large} explored quality disparities using "score gaps" in score-based LLM-as-a-Judge, this factor remains under-explored in comparative settings, which we address by introducing "answer quality gap" for both pairwise and list-wise evaluations.
We define the quality of a candidate's solution by how effectively it addresses the question. Consequently, the answer quality gap refers to the disparity in quality between the solutions from one candidate model and the others to the same question and hence considered the Candidate-level factor. Ideally, when a reliable LLM judge is presented with a question and corresponding answer pairs or series, it would prefer the highest-quality answer, where the corresponding candidate is denoted as the winner selected by the LLM judge. 

Following this assumption, we measure the answer quality gap by the win rates of candidates over an expected baseline on a set of tasks and questions. However, if position bias occurs, the winner may be inconsistent when the order of candidate solutions is permuted in the query. Therefore, we categorize the LLM judgments into three groups: cases where the same winner is consistently chosen across all permutations (termed “consistent wins”), cases where there is no certain winner (termed “consistent ties”), and cases where different winners are selected after the solutions are permuted in the queries (termed ``inconsistent judgment series"). We denote these counts as the number of consistent wins ($C_{w}$), consistent ties ($C_{t}$), and inconsistent judgment series ($C_{I}$), respectively. Inspired by \citeauthor{zheng2024judging}, we count inconsistent judgment pairs as ties for all candidate models, which is later calculated as a down-scaled win rate depending on the number of candidate models.

To calculate the win rates of candidate models for all three cases, we define the \textbf{overall win rate} ($owr$) of a model's solution over the other as: $owr = \frac{1}{n}[{C_w + \frac{1}{p}(C_t + C_I)}]$, where we have $p$ candidates in the list and $n$ judgment series. Then the \textbf{answer quality gap ($\delta_{q}$)} is calculated as $\delta_{q} = |owr - 1/p|$, where $1/p$ is the expected baseline when all judgments are ``ties''. In contrast to using only \textit{consistent win rate} (calculated as $\frac{C_w}{n_c}$, where $n_c$ is the number of position consistent judgment series) to quantify $\delta_{q}$ \citep{zheng2024judging,li2023generative,raina2024llmasajudge}, the adoption of overall win rate incorporates all data points and captures the “comparable quality” cases, where responses in similar quality might lead to position biased judgments, a scenario that the consistent win rate might overlook. 
\section{Experiment}
\subsection{Experiment Settings}

In this study, we evaluated position bias of 15 models from the GPT \citep{openai2024gpt4}, Claude \citep{anthropic2024claude3}, Gemini \citep{geminiteam2024gemini}, and Llama \citep{Llama2023} series using our framework.
For datasets, we adopted the modified MTBench \citep{zheng2024judging} and DevBench \citep{li2024devbench} due to their demonstrated high human-LLM agreement and the validated reliability of state-of-the-art LLM judges on the evaluation tasks. For pairwise comparisons, We fixed one of the candidates as \textbf{vicuna-13b-v1.3} for MTBench and \textbf{human} for DevBench to serve as baselines, ensuring decent quality of solutions to the given questions. MTBench consists of 30 candidate models, 8 tasks, and 10 questions per task; for DevBench, we divide the \textit{general} metric into more detailed ones and consider them as different tasks, resulting in 10 candidate models, 14 tasks, and 8 questions per task. We then paired solutions of these candidate models with that of the baseline candidate for evaluation by the LLM judges. 

We adopted Two-option mode for MTBench, and Three-option mode for DevBench. 
For list-wise experiments, we randomly sampled 9 models to form three triple-candidate lists and evaluated four representative LLM judges on MTBench.
The prompt templates we used are identical to those in the benchmarks for pairwise comparisons, with minor modifications to accommodate list-wise evaluations. More details about the models, tasks, and prompts can be found in Appendix. Sec. \ref{appendix:experiment_settings}. 

To compute repetition stability, we sampled 3 questions per task and 4 candidate models, paired with baseline candidates, for each LLM judge to evaluate across 3 repetitive trials. This resulted in 576 instances per judge for MTBench and 432 instances per judge for DevBench. The \textit{temperature} hyperparameter was set to 1 for all LLM judges to generate nontrivial results. To compute position consistency and preference fairness, the number of instances increased to 4,800 and 2,240, covering the entire MTBench and DevBench datasets. In total, more than 100,000 evaluation instances were analyzed in this study.

To identify significant factors contributing to position bias, we performed bidirectional stepwise regression on data from the two benchmarks. We used variables such as average lengths of input, output, and prompt; answer quality gap; LLM judge series; candidate identities; and task categories to predict $PC$ and $PF$, respectively. Each model prunes non-significant variables based on the Akaike Information Criterion (AIC) score. This process involves both forward selection and backward elimination, with each "step" testing whether including or excluding a variable improves the model's AIC value. Further details about the process can be found in Appendix. Sec. \ref{appendix:variable_select}. 

\begin{table}[ht]
  \centering
    \newcommand{\redbg}[1]{\cellcolor{red!15}#1}

  \scalebox{0.7}{
  \begin{tabular}{lccc}
    \toprule
    Factor & Judge-level & Candidate-level & Task-level  \\
    \midrule
    Familial Property & \redbg{$\checkmark$*} & $\times$ & $\times$ \\
    Answer Quality Gap & $\times$ & \redbg{$\checkmark$*}  & $\times$ \\
    Task Input Length & $\times$ & $\times$ & $\checkmark$ \\
    Task Output Length & $\times$ & $\checkmark$* & $\checkmark$* \\
    Prompt Length & $\times$ & $\checkmark$ & $\checkmark$ \\
    \bottomrule
  \end{tabular}
 }
 \caption{Factors influencing position bias. Significant factors, identified via bidirectional stepwise regression, are marked with * and highlighted in \colorbox{red!15}{red} based on empirical findings on both MTBench and DevBench results. Task Input refers to the question itself, while Task Output denotes the candidate model's answers, serving as both Candidate-level and Task-level factors. Prompt includes the full query presented to LLM judges: Task Input, Task Output, and system prompts.}
  \label{tbl:impact-factors}
  \vspace{-6mm}
\end{table}

\subsection{Empirical Results}

The evaluation results of 12 close-source and 3 open-source LLM judges in terms of repetition stability, position consistency, and preference fairness on MTBench and DevBench are listed in Table \ref{tbl:rs_pc_pf_results}. For each judge, we calculate its average $RS$, $PC$, and $PF$ across all candidates and tasks. For $RS$ and $PC$, higher values are preferable. A high $RS$ value is particularly important as a prerequisite for meaningful computations of $PC$ and $PF$, ensuring the LLM judge's choice patterns are not due to random variations. Fig. \ref{fig:MTBench_main_results} (a)(b) demonstrate that position bias varies by judges and tasks significantly. Fig. \ref{fig:MTBench_main_results} (c) explores the correlation between the metrics $PC$ and $PF$. Fig. (d) to (f) further investigate the impact of the answer quality gap on position bias. These analyses were conducted by considering all judges together on MTBench. More analyses can be found in Appendix. Sec. \ref{appendix:more_results}.

Through bidirectional stepwise regression, as shown in Table \ref{tbl:impact-factors}, LLM judge series, candidate identities, and task categories significantly impact Position Consistency among all variables. Similarly, these factors also contribute significantly to Preference Fairness. Additionally, we found that average output length is a statistically significant predictor of $PF$. This finding is not surprising, as longer outputs are generally perceived as higher quality and more preferred.
Quantitative results and more discussions can be found in Appendix. Sec. \ref{appendix:variable_select}.

\begin{table*}[ht]
    \centering
    \setlength{\tabcolsep}{4pt}
    \scalebox{0.71}{
    \begin{tabular}{l|cccc|cccc|ccc}
    \hline
    \multirow{2}{*}{Judge} & \multicolumn{4}{c|}{MTBench Pairwise} & \multicolumn{4}{c|}{DevBench Pairwise} & \multicolumn{3}{c}{MTBench List-wise} \\ 
    \cline{2-12}
    & $RS$ & $PC$ & $PF$ & Error & $RS$ & $PC$ & $PF$ & Error & $PC$ & $PF$ & Error \\
    \hline
    Claude-3.5-Sonnet & 0.96 $\pm$ 0.07 & \textbf{0.82 $\pm$ 0.14} & \textbf{0.01} & 0.00 & 0.95 $\pm$ 0.09 & 0.76 $\pm$ 0.16 & 0.22 & 0.00 & 0.67 $\pm$ 0.19 & 0.17 $\pm$ 0.19 & 0.00 \\
    Claude-3-Opus & 0.95 $\pm$ 0.08 & 0.70 $\pm$ 0.19 & 0.22 & 0.00 & 0.96 $\pm$ 0.07 & 0.69 $\pm$ 0.20 & 0.29 & 0.00 & \cellcolor{gray!15} & \cellcolor{gray!15} & \cellcolor{gray!15} \\
    Claude-3-Sonnet & 0.93 $\pm$ 0.11 & 0.59 $\pm$ 0.22 & 0.32 & 0.01 & 0.95 $\pm$ 0.09 & 0.71 $\pm$ 0.22 & 0.23 & 0.00 & \cellcolor{gray!15} & \cellcolor{gray!15} & \cellcolor{gray!15} \\
    Claude-3-Haiku & 0.89 $\pm$ 0.18 & 0.57 $\pm$ 0.18 & 0.18 & 0.00 & 0.90 $\pm$ 0.17 & 0.23 $\pm$ 0.14 & 0.75 & 0.00 & \cellcolor{gray!15} & \cellcolor{gray!15} & \cellcolor{gray!15} \\
    
    \addlinespace[0.5em]
    Gemini-1.5-pro & 0.97 $\pm$ 0.09 & 0.62 $\pm$ 0.19 & 0.23 & 0.03 & 0.87 $\pm$ 0.17 & \textbf{0.84 $\pm$ 0.17} & \textbf{0.03} & 0.13 & 0.55 $\pm$ 0.20 & 0.33 $\pm$ 0.18 & 0.00 \\
    Gemini-1.0-pro & 0.89 $\pm$ 0.18 & 0.57 $\pm$ 0.18 & 0.30 & 0.00 & 0.85 $\pm$ 0.26 & 0.66 $\pm$ 0.20 & \textbf{-0.05} & 0.00 & \cellcolor{gray!15} & \cellcolor{gray!15} & \cellcolor{gray!15} \\
    Gemini-1.5-flash & 1.00 $\pm$ 0.00 & 0.67 $\pm$ 0.17 & 0.07 & 0.00 & \cellcolor{gray!15}\textcolor{red}{0.04 $\pm$ 0.08} & \cellcolor{gray!15}0.92 $\pm$ 0.39 & \cellcolor{gray!15}0.00 & \cellcolor{gray!15}\textcolor{red}{0.96} & \cellcolor{gray!15} & \cellcolor{gray!15} & \cellcolor{gray!15} \\

    \addlinespace[0.5em]
    GPT-4 & 0.97 $\pm$ 0.05 & \textbf{0.82 $\pm$ 0.15} & \textbf{0.02} & 0.00 & 0.97 $\pm$ 0.05 & \textbf{0.83 $\pm$ 0.15} & -0.13 & 0.00 & \cellcolor{gray!15} & \cellcolor{gray!15} & \cellcolor{gray!15} \\
    GPT-4-Turbo & 0.94 $\pm$ 0.10 & 0.75 $\pm$ 0.16 & \textbf{0.02} & 0.00 & 0.97 $\pm$ 0.06 & 0.79 $\pm$ 0.18 & 0.16 & 0.00 & \cellcolor{gray!15} & \cellcolor{gray!15} & \cellcolor{gray!15} \\
    GPT-4o & 1.00 $\pm$ 0.02 & 0.76 $\pm$ 0.18 & -0.12 & 0.00 & 0.98 $\pm$ 0.03 & \textbf{0.80 $\pm$ 0.16} & -0.12 & 0.00 & 0.68 $\pm$ 0.22 & 0.18 $\pm$ 0.22 & 0.00 \\
    GPT-3.5-Turbo & 0.96 $\pm$ 0.07 & 0.70 $\pm$ 0.18 & 0.06 & 0.00 & 0.99 $\pm$ 0.02 & 0.76 $\pm$ 0.18 & \textbf{-0.02} & 0.00 & 0.34 $\pm$ 0.17 & -0.05 $\pm$ 0.30 & 0.12 \\
    o1-mini & 0.90 $\pm$ 0.07 & \textbf{0.76 $\pm$ 0.15} & \textbf{-0.04} & 0.00 & 0.93 $\pm$ 0.12 & \textbf{0.84 $\pm$ 0.13} & -0.07 & 0.00 & \cellcolor{gray!15} & \cellcolor{gray!15} & \cellcolor{gray!15} \\

    \addlinespace[0.5em]
    Llama-3.3-70B & 0.96 $\pm$ 0.06 & \textbf{0.80 $\pm$ 0.16} & \textbf{-0.05} & 0.00 & 0.99 $\pm$ 0.01 & \textbf{0.89 $\pm$ 0.12} & \textbf{-0.03} & 0.00 & \cellcolor{gray!15} & \cellcolor{gray!15} & \cellcolor{gray!15} \\
    Llama-3.1-405B & 0.93 $\pm$ 0.10 & \textbf{0.77 $\pm$ 0.16} & 0.10 & 0.02 & 0.94 $\pm$ 0.10 & 0.79 $\pm$ 0.15 & \textbf{0.01} & 0.00 & \cellcolor{gray!15} & \cellcolor{gray!15} & \cellcolor{gray!15} \\
    Llama-3.1-8B & 0.75 $\pm$ 0.32 & 0.69 $\pm$ 0.23 & -0.03 & 0.25 & 0.79 $\pm$ 0.36 & 0.47 $\pm$ 0.18 & 0.25 & 0.00 & \cellcolor{gray!15} & \cellcolor{gray!15} & \cellcolor{gray!15} \\
    \hline

    \end{tabular}
    }
    \caption{Evaluation results for Repetition Stability ($RS$), Position Consistency ($PC$), and Preference Fairness ($PF$) are presented for both pairwise and list-wise evaluation approaches, with the top 5 performances marked in \textbf{bold}. \textbf{Errors} arise from judgment failures (e.g., exceeding context window, not following output format). High error rates and low $RS$ are marked {\color{red} red}, rendering further evaluations \colorbox{gray!15}{invalid} due to insufficient data. List-wise evaluation is conducted on four representative judges to validate scalability.}
    \label{tbl:rs_pc_pf_results}
    \vspace{-6mm}
\end{table*}

\section{Main Findings}

\begin{figure*}[ht]
    \centering
    \includegraphics[scale=0.25]{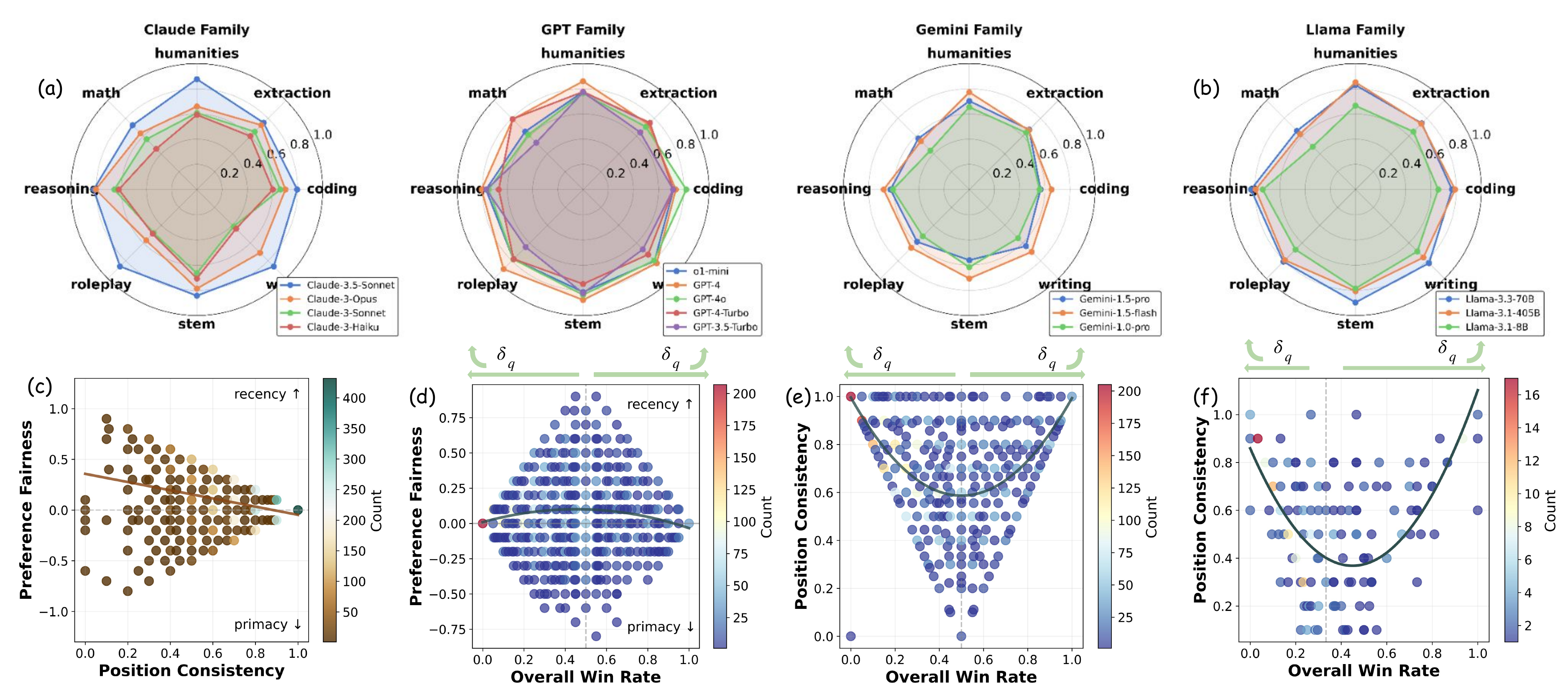}
    \caption{Judge performances on MTBench. Fig. (a)(b) are the radar charts for the $PC$ comparison by family, judge, and task. Fig. (c) leverages linear regression to explore the general relationship between $PC$ and $PF$. Fig. (d) to (f) investigate the impact of answer quality gap on position bias using overall win rates. Fig. (a) to (e) are for pairwise comparative settings, while Fig. (f) are obtained under list-wise evaluations.}
    \label{fig:MTBench_main_results}
    \vspace{-3mm}
\end{figure*}

\textbf{Position Bias of Capable Judges are not Mere Random Variations:} As shown in Table \ref{tbl:rs_pc_pf_results}, the capable judges on the benchmark tasks, supported by minimal "Error" rates, generally exhibit $RS$ values above 0.85. The most capable models, such as Claude-3.5-Sonnet, GPT-4, and Llama-3.3-70B, all achieve near-perfect $RS$ scores over 0.95 on both benchmarks. These results confirm that judgments from capable LLM judges, and the resulting position bias, are not due to random variations. This strengthens confidence that one-time generated judgments by these validated LLMs accurately reflect their judging capabilities.

\textbf{Position Bias Varies by Judge \& Task:}
As shown in Fig. \ref{fig:MTBench_main_results}(a), position bias among LLM judges varies significantly across different judges and tasks. For instance, GPT-4o demonstrates higher position consistency when evaluating coding tasks but performs less consistently on other tasks compared to GPT-4. Similarly, Gemini-1.5-pro, while achieving higher $PC$ than other Gemini models in most tasks, exhibits comparable consistency when judging extraction tasks. Similar findings can be observed in the DevBench results, as detailed in Appendix.Sec. \ref{appendix:DevBench}.

Variations in preference fairness are also evident. As shown in Table \ref{tbl:rs_pc_pf_results}, GPT-4 and GPT-3.5-Turbo display different preference biases across datasets and tasks—being recency-preferred on MTBench but primacy-preferred on DevBench. Likewise, Claude-3.5-Sonnet, which is nearly preference-fair on MTBench ($PF = 0.01$), exhibits a strong recency-preferred position bias on DevBench ($PF = 0.22$).

While higher position consistency generally correlates with improved preference fairness (as demonstrated by the regression curve in Fig. \ref{fig:MTBench_main_results}(c)), consistency alone does not guarantee fairness. Certain LLM judges, despite achieving high $PC$, still exhibit significant and varied preference directions across different tasks, underscoring the need to evaluate both consistency and fairness when assessing LLM judges.




In list-wise comparisons, similar variations in position bias were observed across judges and tasks. Furthermore, Table \ref{tbl:rs_pc_pf_results} highlights that more capable models, such as GPT-4o and Claude-3.5-Sonnet, maintain high consistency when transitioning from pairwise to list-wise evaluations, while less capable models, such as GPT-3.5-Turbo, exhibit greater sensitivity to the increased number of candidates in list-wise tasks.

Therefore, the position bias of LLM judges is both judge-dependent and task-dependent. This observation is further confirmed by the bidirectional stepwise regression where judge identities and task categories are statistically significant predictors of $PC$ and $PF$. In practice, when evaluating LLM judge's reliability or selecting suitable LLM judges, considering the balance between consistency and fairness, as well as accounting for task-specific variations, may be beneficial.

\textbf{Position Bias Correlates to Answer Quality Gap:} Intuitively, the difficulty of judging a pair of candidate answers is largely reflected by their difference in quality. In this study, as defined in Section \ref{sec:factors_affecting_position_bias}, we quantify the quality gap ($\delta_q$) between candidate solutions and expected baseline (calculated by $1/p$ for a $p$-candidate list) by the overall win rate ($owr$). Therefore, $\delta_q$ increases as $owr$ extends from baseline to 0 or 1. Fig. \ref{fig:MTBench_main_results} (e) and (f) exhibit significant parabolic shapes, indicating that $PC$ is positively proportional to $\delta_q$. This aligns with our intuition that the answer pairs or series with larger quality disparities are easier to achieve judgment consistency, whereas those of similar quality are difficult to judge, increasing the likelihood of position bias that leads to lower $PC$. The same relationship is observed for each individual judge and across benchmarks, as demonstrated in Appendix.Sec. \ref{appendix:more_results}.

Similarly, as shown in Fig. \ref{fig:MTBench_main_results} (d), judgments generally become more preference fair as $\delta_q$ increases. However, the extent is not as significant as for $PC$. Also, the relationship varies by judge, as some LLMs maintain preference fairness regardless of $\delta_q$. For example, as shown in Appendix.Fig. \ref{fig:MTBench_winrate_each_judge}, $PF$ of GPT models centered closely around 0 consistently, whereas that of Claude and Gemini-pro models exhibit a conspicuous proportional relationship on MTBench. These observations align with the right-arrow shape as demonstrated in Fig. \ref{fig:MTBench_main_results} (c), where there is a general trend that judgments become preference fairer as position consistency increases. It also justifies the reasonableness of our quantification of preference fairness, as highly position consistent judges are not overly penalized and a perfect $PC$ should result in $PF=0$.

Together, we conclude that as the answer quality gap enlarges, judges \textbf{generally} become more position consistent and preference fair according to the regression curves. However, exceptions are common, as shown by the individual scatter points of these figures. This indicates that though the answer quality gap significantly influences the position bias of LLM judges, other factors also play important roles. Therefore, built on our findings, future studies may have better control over the answer quality gap when evaluating LLM judges, exploring other impacting factors on position bias, and seeking potential mitigation strategies.

\textbf{Position Bias is weakly Length-dependent} We investigate the impact of three different lengths on the position bias of LLM judges: the length of the question (task input length), the solution length of candidate models (task output length), and the length of the entire prompt (prompt length). By stepwise regression, we discovered that average task output length is only significant in predicting $PF$, adding a minimal change in AIC as shown in Appendix Table. \ref{tab:both_benchmarks_PF_OLS}. In other words, there is a very weak relationship between the lengths of prompt components and position bias. 



\textbf{LLM Agreement Analysis:} We complement our investigation of position bias with an agreement and disagreement analysis among LLM judges. Rather than focusing exclusively on overall consistency or fairness, we examine how LLM judges converge and diverge in their assessments of individual instances. Agreement analysis quantifies the percentage of instances where two LLM judges mutually agree on the outcome. Disagreement analysis counts the number of choices deviating from the mode for each instance among all judges. This further complies a ``distribution of disagreement'' across the dataset. 

\begin{figure}[ht]
    \centering
    \includegraphics[scale=0.45]{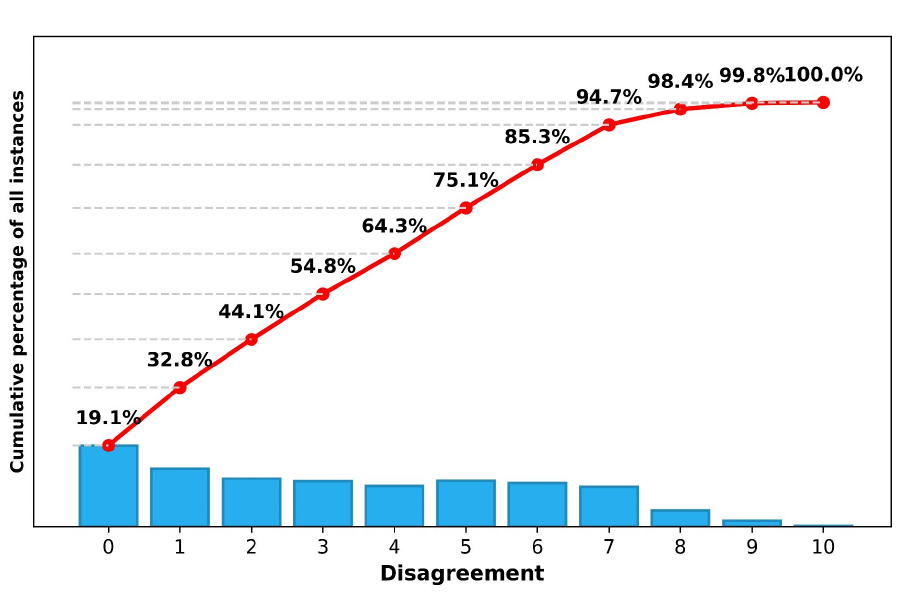}
    \caption{Distribution of disagreement on MTBench. The y-axis indicates the proportion of the dataset where the level of disagreement among LLM judges does not exceed a specific threshold.}
    \label{fig:MTBench_disagreement}
    \vspace{-6mm}
\end{figure}

Our findings reveal that, despite exhibiting similar overall $PC$ and $PF$ scores, judges vary significantly in their judgments on individual instance. Disagreement analysis, in particular, highlights instances where consensus is either easily or difficultly achieved, reflecting the inherent complexity of the judgment task. 
For example, as shown in Fig. \ref{fig:MTBench_disagreement}, more than half of the dataset can be considered relatively easy to judge, as over 80\% of all 15 LLM judges agree with each other on these instances (disagreement $\leq$ 3). Conversely, fewer than 2\% of instances represent the likely especially-hard-to-judge cases where a majority of LLM judges fail to reach consensus (disagreement $\geq$ 8).

Based on our observations of answer quality gaps and LLM agreement/disagreement patterns, this study offers practical insights for designing evaluator benchmarks that account for the varying difficulty levels of judgment tasks. Specifically, the most challenging instances to evaluate are characterized by: (1) frequent disagreements among LLM judges, (2) closely matched win rates and minimal quality gaps among candidate models, and (3) significant position bias exhibited by the majority of LLM judges. Further discussions and analyses can be found in Appendix Sec. \ref{appendix:agreement_analysis}.

\section{Conclusion}

In conclusion, this paper provides an in-depth analysis of position bias in LLM judges, a critical challenge in automated evaluation. Using metrics such as repetition stability, position consistency, and preference fairness, we identify significant variations in position bias across judges and tasks, consistent across pairwise and list-wise comparison settings. Our findings show that position bias is weakly influenced by prompt length but strongly impacted by the quality gap between solutions. Furthermore, agreement and disagreement analysis highlights variability in judgment reliability, providing valuable insights for refining benchmarks. This study enhances understanding of position bias and contributes to the development of fairer and more reliable LLM evaluation frameworks.


\clearpage
\section{Limitations}
\label{appendix:limitation}
Despite proposing scalable metrics and investigating key factors influencing position bias, our study has several limitations.

First, we evaluated only 12 commercial closed-source LLM judges for pairwise settings and 4 for list-wise paradigms across two benchmarks, limiting the list-wise comparisons to three-candidate lists. Among open-source models, we assessed only three Llama 3.1+ models of varying sizes, as earlier versions or other smaller models lacked sufficient context window lengths, making them unsuitable given the length of evaluation instances in our study. Additionally, while we used original benchmark prompt templates, exploring alternative prompting techniques could offer further insights. Future work could expand on this by incorporating more models, tasks, prompting strategies, and larger list-wise candidate pools to enhance the generalizability of our findings.

Second, data accessibility limitations prevented a direct analysis of Judge-level factors like architecture and parameter size of closed-source models. Instead, we approximated these factors by grouping models by family properties. While open-source models offer accessible architectural details for deeper analysis, our assessment of only three Llama models may not provide sufficient evidence for broader conclusions. Additionally, our analyses were conducted post hoc, relying on completed judgments before analysis. Future work could explore methods to estimate or control these factors pre-judgment, reducing computational costs and enabling proactive mitigation strategies.

Lastly, our focus was on evaluating and understanding position bias rather than mitigating it. While our findings provide a foundation for effective mitigation, further research is needed to address issues like maintaining consistency and fairness when answer quality gaps are minimal, where position bias is most pronounced. Multivariate analyses exploring interactions between factors like prompt length, task complexity, and answer quality gaps could also yield deeper insights and enhance mitigation approaches.

\bibliography{reference}

\clearpage
\appendix
\startcontents[appendices]
\section*{Appendix Table of Contents}
\printcontents[appendices]{}{0}{\large}
\newpage

\section{Reproducibility}
Our experiments were conducted primarily using API access, with a total cost of approximately 3,000 USD. The code repository for reproducing our results is available:\url{https://anonymous.4open.science/r/Position-Bias-Analyzer-Demo-F7E3}
\section{Related Work}

\subsection{LLM-as-a-Judge}
Large Language Models (LLMs) have become a transformative tool in automating evaluative tasks, offering scalability and reproducibility advantages over human assessments \citep{zheng2024judging}. The methodology of using LLMs as evaluators ("LLM-as-a-Judge") has been widely used for tasks such as open-ended story generation \citep{chiang-lee-2023-large}, adversarial attacks \citep{chiang2023can}, summarization \citep{karpinska2021perils}, machine translation \citep{kocmi-federmann-2023-large}, and instruction following \citep{zeng2023evaluating}, where models are tasked with scoring or ranking outputs. LLM-as-a-Judge paradigms range from pointwise, pairwise, and list-wise ranking approaches \citep{qin2024large, zhu2024starlingb}
to score-based and relation-based (comparative or ranking) settings \citep{li2024split}. Furthermore, explorations into multi-agent frameworks \citep{li2023prd,chan2023chatevalbetterllmbasedevaluators}
and meta-thinking or meta-judge systems \citep{wu2024metarewardinglanguagemodelsselfimproving,li2025leveragingllmsmetajudgesmultiagent}
have emerged to enhance reliability of LLM evaluations. Despite their potential, inherent biases—particularly position bias—pose significant challenges to their reliability and fairness \citep{zheng2024judging,ye2024justiceprejudicequantifyingbiases,ma2025judging}, even in the most effective pairwise comparative settings.

\subsection{Position Bias}

Position bias, the tendency of LLMs to favor certain positions within the prompt irrespective of actual content, is a pervasive issue observed across various domains and tasks \citep{zheng2024judging, qin2024large, li2024devbench,li2024split}. In the context of LLM-as-a-Judge, it typically refers to the position inconsistent scenarios where a LLM judge prefers different candidate solutions to a given question when the order of solutions in the prompt is permuted. To deal with such bias, the naive way is to exclude the position inconsistent judgments, which does not solve the fundamental issue and would likely result in data sparsity when position bias is frequently exhibited. Then, researchers proposed `inconsistency-as-a-tie' for both candidate models in pairwise comparative settings to consider all judgments for further analysis \citep{zheng2024judging,li2023prd}. This approach, while practically useful for evaluations, does not mitigate position bias.

Given the significance of position bias, researchers have developed debiasing strategies, including bootstrapping \citep{hou2024large}, split-and-merge techniques \citep{li2023split}, and multi-agent discussions \citep{li2023prd,khan2024debating}. While these methods demonstrate potential, they are often costly, time-consuming, or insufficient to eliminate position bias. More recently, mechanistic approaches have been explored to address position bias by modifying model internals like positional embeddings and causal attention mechanisms \citep{yu2025mitigate,wang2025eliminating}. These methods, while effective in eliminating position bias, lack applicability to state-of-the-art closed-source models, which are frequently used in practice but model internals are inaccessible. Therefore, position bias remains a complex and challenging issue to fully resolve.

\subsection{Evaluating Position Bias Evaluation}

Parallel to developing debiasing strategies, a crucial line of research focuses on comprehensively assessing position bias and identifying its influencing factors, thereby fostering a deeper understanding of this phenomenon. The evaluation of position bias typically concentrates on two main aspects: the \textit{consistency} of judgments when the order of candidate solutions in the prompt is changed, and the \textit{preference direction} (e.g., primacy or recency bias) when inconsistencies arise. Metrics for these aspects in pairwise comparisons are relatively established; however, their extension to, and rigorous measurement in, list-wise settings remain less explored.

While prior works \citep{qin2024large,zhu2024starlingb} introduced list-wise ranking settings, they often become computationally expensive due to exponentially increasing number of pairs in the list or encounter significant degradation of performance due to ranking complexity. Moreover, the traditional preference direction measurement of position bias becomes increasingly complicated in list-wise paradigms (i.e., preferring the \textit{$i^{th}$} candidate solution for all \textit{i}'s). However, practically, ``choosing the best out of an n-item list'' is a simpler but common need compared to exhaustive ranking of the entire list, which is not yet fully explored by prior works. A unified single metric for effectively measuring preference direction is therefore necessary to facilitate position bias evaluation in a more comprehensive manner.

\subsection{Factors influencing Position Bias}
Position bias may be influenced by several factors, including judge attributes, task types, candidate solution lengths and qualities, and more. For example, \cite{wang2023large} discovered that judgment conflict rate (i.e., frequency where position bias exhibits across the dataset) is negatively correlated with the score gap between the two candidate responses. However, quantifying and understanding the impact of this ``answer quality gap" in comparative settings (both pairwise and list-wise) is an area that requires further investigation.

Besides, prior works often observe position bias and conduct subsequent analyses based on single-instances judgments. However, observed biases might stem from random variations rather than systematic tendencies, hence compromising the reliability of downstream bias analyses. To mitigate this, researchers have employed multiple repetitions and used the mode judgment, an approach that, while robust, can be computationally expensive \citep{chen2024mllmasajudge}. In this sense, validating that position bias emerges from positional information rather than randomness becomes a crucial yet insufficiently-explored area.

These identified gaps - spanning from the validation of judgments to the evaluation of list-wise settings - highlight the need for a more comprehensive framework to evaluate and understand position bias. By addressing these dimensions collectively, our work provides a foundation for a deeper understanding of the consistency, fairness, and reliability of LLM evaluators.
\section{LLM Agreement Analysis}
\label{appendix:agreement_analysis}
Besides the exploration of position bias with a broad lens by average $PC$ and $PF$, instance-wise agreement between LLM judges is also insightful. Even two judges with the same $PC$ and $PF$ scores may not reach consensus on each instance. Therefore, this session investigates (1) what percentage of a set of evaluations do two LLM judges agree on each other? (2) how do the choices of all judges on an instance vary?

\begin{figure*}[ht]
    \centering
    \includegraphics[width=0.9\textwidth]{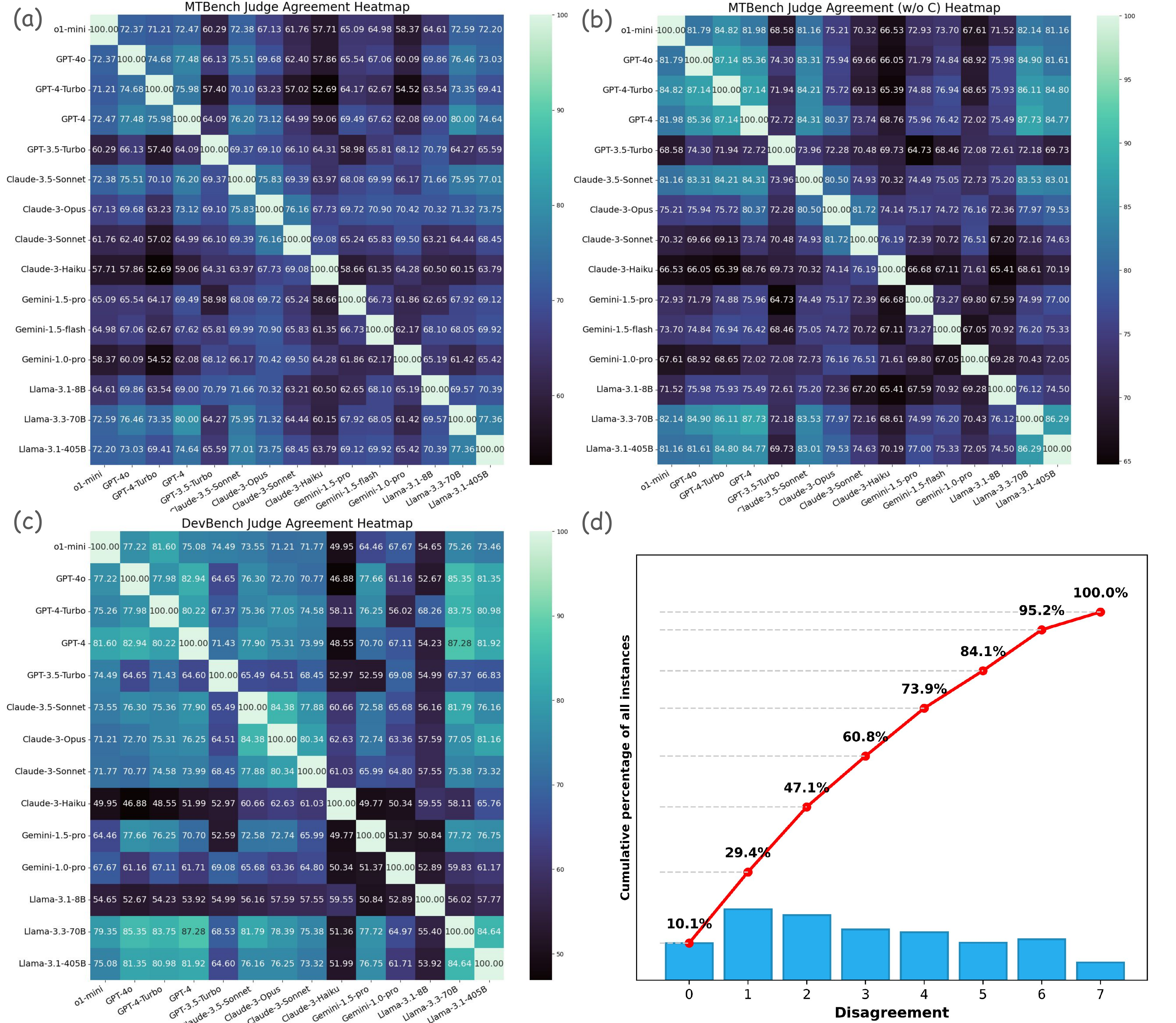}
    \caption{Figures (a) to (c) are mutual agreement heatmap of LLM judges on MTBench and DevBench, where (b) is the agreement computation excluding the tie option \{C\}. Higher mutual agreement between two LLM judges is marked with brighter color. Figure (d), like Figure \ref{fig:MTBench_disagreement}, is the distribution of disagreement on DevBench.}
    \label{fig:agreement_analysis}
\end{figure*}

\subsection{Mutual Agreement \& Familial Property}
We compute the LLM judges' mutual agreement on the instances to explore how ``alike'' or consistent they are across a set of evaluations. We denote two judges \textit{agree} on an instance if their judgment choices are identical. Then the \textit{mutual agreement} between two LLM judges on a benchmark is defined as the proportion of their agreed instances. Figs. \ref{fig:agreement_analysis}(a) and (c) displays the mutual agreement heatmap for all judges on MTBench and DevBench, respectively. For MTBench that utilizes the 3-option mode, we also consider the ``without tie" agreement since two judges are less disagreed when one chooses \{C\} while the other prefers a certain solution, compared to the case when they prefer different solutions. The ``without tie'' agreement heatmap of the twelve judges on MTBench is explored in Fig. \ref{fig:agreement_analysis}(b).

The heatmaps reveal clear ``familial patterns'' in the judgment choices of these LLM judges. For instance, the GPT-4, GPT-4-Turbo, and GPT-4o series exhibit high agreement on MTBench, achieving over 70\% with ties included and over 85\% without. GPT-3.5-Turbo didn't agree with the GPT-4 series and o1-mini for around 40\% of the instances, indicating that they are considerably different in judging capabilities. 

For Claude-3 models, similar familial patterns could be observed. Claude-3-Opus highly agrees with Claude-3.5-Sonnet, probably due to their similar capabilities, while it also highly agrees with Claude-3-Sonnet, likely due to their similar model structure within the same series. Interestingly, Claude-3.5-Sonnet and Claude-3-Sonnet do not exhibit a significantly high agreement, indicating that the upgrade from series 3 to 3.5 considerably impacts their judging capabilities. 

Gemini models exhibit rather low mutual agreement and ``familial property" is minimal, but the most capable Gemini-1.5-pro aligns more closely with other capable models like the GPT-4 series and Claude-3-Opus. 

Llama models demonstrate a high agreement among capable family members (Llama-3.3-70B and Llama-405B) and with GPT series. However, significantly smaller and less capable model like Llama-3.1-8B does not strong agree with them.

These patterns suggest that familial similarities, possibly stemming from analogous model sizes, training data, and strategies, influence the positional preferences of these judges. In particular, the LLM judges could be primarily grouped by their capabilities; when judging capabilities are comparable, models within the same family series share a higher mutual agreement than across families.

Identifying such groupings provides valuable insights, as comparisons between judges from different groups, both adept at assessing LLM-generated content, can reveal distinct position biases and enrich our understanding of this phenomenon.

\subsection{Disagreement \& Benchmark Design Insight}
Since the mutual agreement between LLM judges is not perfect and usually a considerable proportion of instances are difficult for them to reach a consensus, disagreement analysis becomes crucial and insightful. Therefore, we define the \textit{disagreement} of an evaluation instance to be the number of judgments different from the majority. By this definition, an instance with all judges reaching a consensus on the better solution will have a disagreement of 0; in contrast, an instance where judgments are widely varied will result in a high disagreement. For our study where 15 judges are investigated, the maximum disagreement of an MTBench instance is 10, accounting for the 5\{A\}-5\{B\}-5\{C\} choice pattern by 3-option mode. On the other hand, for DevBench instances, since Gemini-1.5-flash is excluded due to insufficient data (as shown by high error rates in Tabel \ref{tbl:rs_pc_pf_results}), the maximum possible disagreement for the remaining 14 judges is 7, representing the 7\{A\}-7\{B\} judgment distribution for the 2-option mode.

The distributions of instances with different disagreement values on MTBench and DevBench are shown in Fig. \ref{fig:MTBench_disagreement} and Fig. \ref{fig:agreement_analysis}(d), respectively. From our disagreement analysis, at least 75\% of the judges reached a choice consensus on more than half of the instances on both benchmarks. These are likely easy-to-evaluate instances, and the reliability of LLM judgments is enhanced by majority voting. In comparison, the instances with the highest disagreement are likely the ones that are difficult to evaluate and where the position bias is most likely to occur. However, luckily, these instances are rare, occupying only less than 5\% for both benchmarks respectively. In other words, majority voting of multiple capable LLM judges could be practically useful for over 95\% of evaluation instances on both benchmarks.

Moreover, if we roughly consider the disagreement value of instances as their difficulty for judging, then Fig. \ref{fig:MTBench_disagreement} and Fig. \ref{fig:agreement_analysis}(d) exhibit a balanced distribution of instances with varied difficulty. This is because, except for the instances with the highest disagreement, the numbers of other instances with varied disagreement do not vary significantly, indicating a smoothly increasing difficulty curve across the benchmark datasets.

To summarize, the practical implications of the disagreement analysis are three-fold. First, it helps identify the instances that are difficult or trivial to judge, benefiting benchmark designs to control the difficulty of evaluation by managing the number of these instances across the dataset. Second, it assists in filtering out instances where majority voting of LLM evaluators are likely to offer reliable judgments without direct comparison with human-annotated evaluations, enhancing the scalability of LLM judges especially when human evaluations are costly. In other words, if one-shot judgments from only one LLM judge are not enough reliable, multiple capable LLMs and the majoring voting strategy could be employed to make the evaluation more convincing. Last but not least, disagreement analysis provides a convenient way to make the difficulty variance of instances varied across the dataset tangible. Since the difficulty of an evaluation instance is closely related to the quality gap between the two solutions and hence position bias, the investigation of the instances where most judges particularly disagree with one another could provide more insights and inspiration for future benchmark designs and potential mitigation strategies for position bias.
\section{More Results of Position Bias and Answer Quality Gap Measurement}
\label{appendix:more_results}

Table \ref{tbl:mtbench_devbench_more_results} presents additional evaluation results for more open-source models, reinforcing our earlier observations. The scores consistently show that position bias varies across judges and tasks, while high repetition stability confirms that such biases are systematic rather than random. These extended results further validate our evaluation findings.

\begin{table*}[ht]
    \centering
    \setlength{\tabcolsep}{4pt}
    \scalebox{0.8}{
    \begin{tabular}{l|cccc|cccc}
    \hline
    \multirow{2}{*}{Model} & \multicolumn{4}{c|}{MTBench Results} & \multicolumn{4}{c}{DevBench Results} \\
    \cline{2-9}
    & $PC$ & $PF$ & $RS$ & Error & $PC$ & $PF$ & $RS$ & Error \\
    \hline
    GPT-OSS-20B & 0.79 $\pm$ 0.16 & -0.04 $\pm$ 0.13 & 0.89 $\pm$ 0.17 & 0.06 & 0.84 $\pm$ 0.16 & -0.04 $\pm$ 0.14 & 0.81 $\pm$ 0.25 & 0.08 \\
    Gemma-3n-E4B & 0.68 $\pm$ 0.17 & -0.07 $\pm$ 0.18 & 0.90 $\pm$ 0.17 & 0.00 & 0.41 $\pm$ 0.24 & 0.40 $\pm$ 0.35 & 0.93 $\pm$ 0.12 & 0.08 \\
    Llama-4-Scout-17B & 0.79 $\pm$ 0.16 & 0.05 $\pm$ 0.15 & 0.95 $\pm$ 0.08 & 0.00 & 0.12 $\pm$ 0.18 & 0.88 $\pm$ 0.18 & 0.98 $\pm$ 0.03 & 0.00 \\
    Mistral-Small-24B & 0.72 $\pm$ 0.19 & -0.02 $\pm$ 0.17 & 0.87 $\pm$ 0.20 & 0.05 & 0.57 $\pm$ 0.18 & 0.27 $\pm$ 0.19 & 0.82 $\pm$ 0.31 & 0.02 \\
    Qwen2.5-72B & 0.77 $\pm$ 0.17 & -0.01 $\pm$ 0.16 & 0.94 $\pm$ 0.09 & 0.00 & 0.81 $\pm$ 0.15 & 0.11 $\pm$ 0.16 & 0.93 $\pm$ 0.13 & 0.00 \\
    Qwen2.5-Coder-32B & 0.74 $\pm$ 0.18 & 0.01 $\pm$ 0.17 & 0.91 $\pm$ 0.14 & 0.00 & 0.81 $\pm$ 0.16 & 0.09 $\pm$ 0.17 & 0.92 $\pm$ 0.13 & 0.03 \\
    Qwen2.5-7B & 0.55 $\pm$ 0.18 & -0.21 $\pm$ 0.23 & 0.87 $\pm$ 0.22 & 0.00 & 0.54 $\pm$ 0.21 & -0.45 $\pm$ 0.21 & 0.93 $\pm$ 0.12 & 0.00 \\
    \hline
    \end{tabular}
    }
    \caption{More evaluation results for Position Consistency ($PC$), Preference Fairness ($PF$), and Repetition Stability ($RS$) across MTBench and DevBench datasets.}
    \label{tbl:mtbench_devbench_more_results}

\end{table*}

\subsection{MTBench}
As shown in Fig. \ref{fig:MTBench_main_results}(c), considering all judges together, a larger answer quality gap generally leads to better position consistency and preference fairness. In this session, we explore whether the discovery is consistent for each individual judge. Same as Section \ref{sec:factors_affecting_position_bias}, we apply the overall win rate to reflect the answer quality gap for visualization.

\begin{figure*}[ht]
    \centering
    \includegraphics[width=0.9\textwidth]{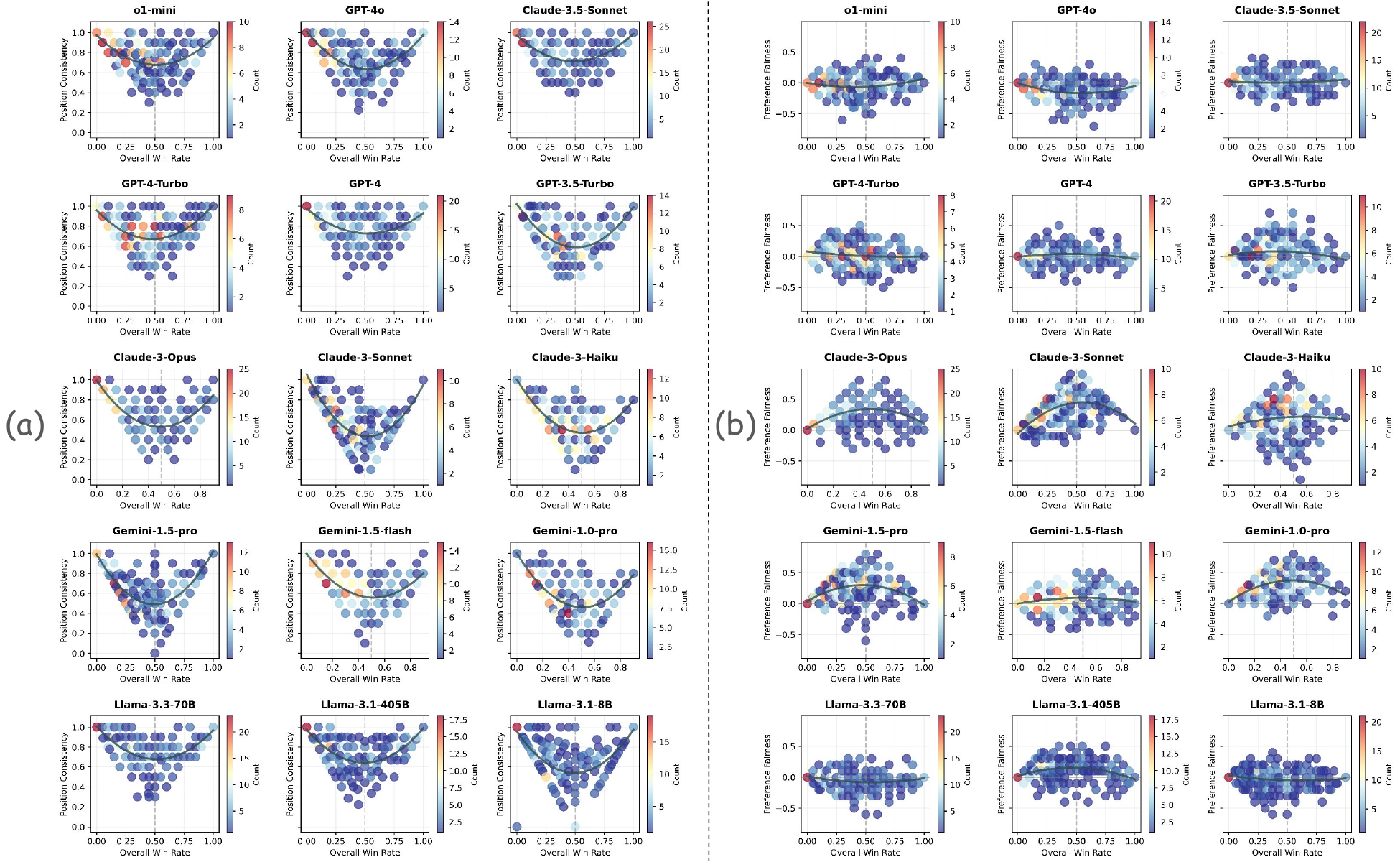}
    \caption{Position Consistency and Preference Fairness vs. overall win rate for each judge on MTBench. Figure (a) refers to the relationship investigation of $PC$ and figure (b) for $PF$.}
    \label{fig:MTBench_winrate_each_judge}
\end{figure*}

As shown in Fig. \ref{fig:MTBench_winrate_each_judge} (a), the ``parabolic shape" is observed for all individual judges, indicating that the argument ``a higher answer quality gap generally results in higher position consistency applies to all models. However, Fig. \ref{fig:MTBench_winrate_each_judge} (b) reveals that preference fairness is more judge-dependent and the impact of the answer quality gap is neglectable for certain judges. For example, while Claude-3-opus and Claude-3-sonnet exhibit conspicuous ``parabolic shape", GPT-4 and GPT-3.5 present nearly linear curves. In other words, while the former models align with the general tendency that a larger answer quality gap improves preference fairness, the latter ones preserve fairness regardless of the answer quality gap. This further demonstrates the necessity to investigate preference fairness in addition to consistency when evaluating a judge model's position bias.

\subsection{DevBench}
\label{appendix:DevBench}
\begin{figure*}[ht]
    \centering
    \includegraphics[width=0.95\textwidth]
    {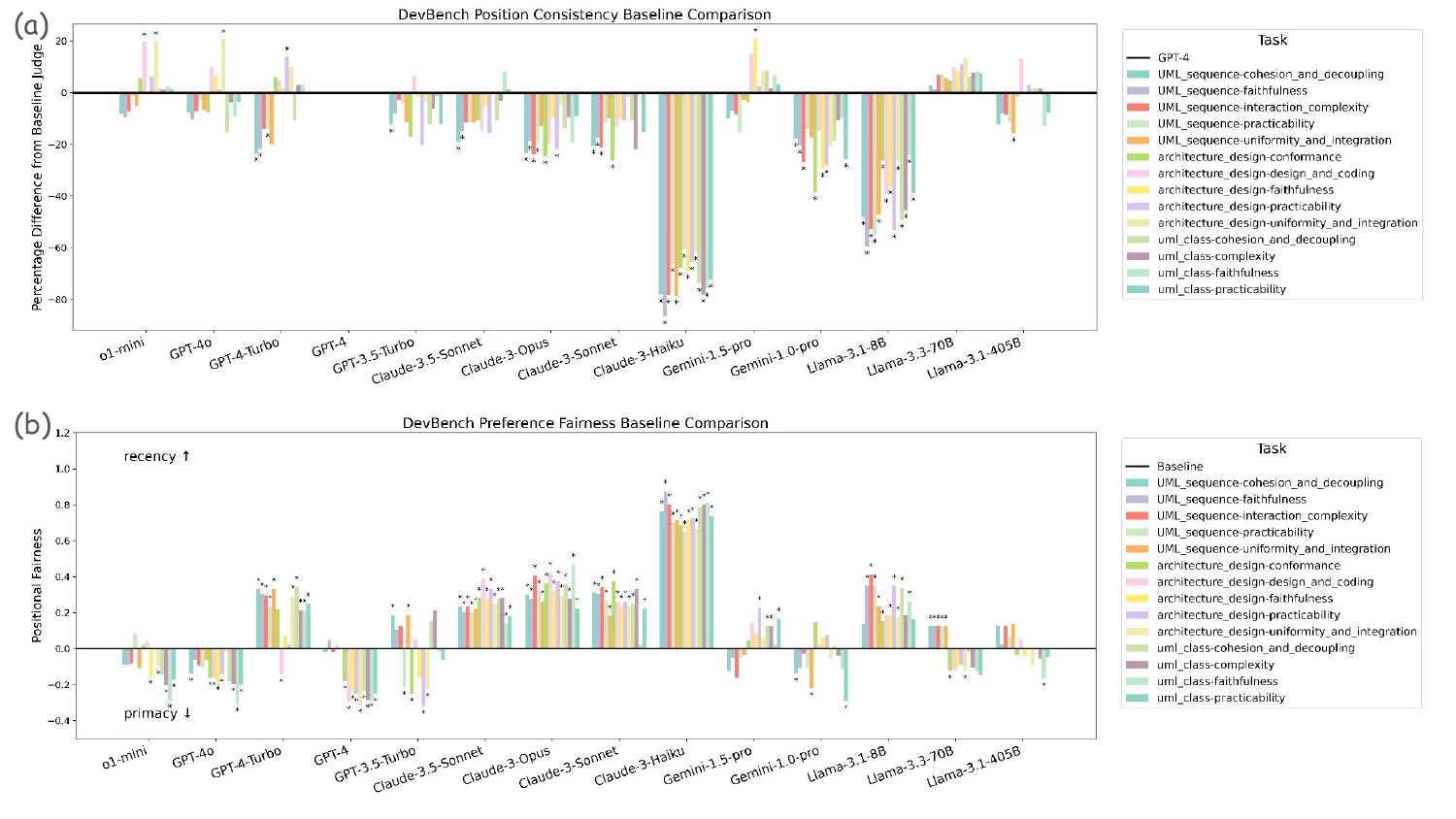}
    \caption{Baseline comparisons of position bias of LLM judges across tasks on DevBench. An asterisk marks the statistical significance by Student's t-tests. Figure (a) selects GPT-4 as the baseline, where asterisk demonstrates signficantly better or worse $PC$ of other models compared to it. Figure (b) utilizes an absolute $PF$ baseline of 0 and depicts preference fairness performances of LLM judges across tasks. Similar to findings on MTBench, position bias is significantly judge-dependent and task-dependent on DevBench as well.}
    \label{fig:DevBench_PC_PF}
\end{figure*}

This session includes a similar baseline comparison analysis on DevBench as on MTBench. As shown in Fig.\ref{fig:DevBench_PC_PF}, position bias is judge-dependent and task-dependent on DevBench as well, as $PC$ and $PF$ vary significantly across judges and tasks. Similarly, although GPT-4 stands as the baseline model with a generally high $PC$ across tasks, certain models achieve comparable or superior performances on certain tasks. For instance, for \textit{architecture design} evaluations, GPT-4-Turbo, GPT-4o, and Gemini-1.5-pro all surpass GPT-4. Gemini-1.5-pro is especially outstanding, also exceeding GPT-4 in \textit{uml class} evaluations. However, GPT-4 is still the best-performing model on \textit{UML sequence} evaluations, with only GPT-3.5-Turbo can achieve comparable performance regarding certain detailed metrics (e.g., interaction complexity). These discoveries, aligning with the findings on MTBench, further necessitate the need to consider the trade-offs between positional consistency and fairness when selecting the optimal judge model for certain tasks.

\section{Variable Selection and Tests}
\label{appendix:variable_select}

\subsection{Bidirectional Stepwise Regression with AIC}

Bidirectional stepwise regression is a combination of forward selection and backward elimination techniques. It iteratively refines the model by adding or removing predictors based on a statistical criterion—commonly the Akaike Information Criterion (AIC). The objective is to select a model that balances goodness of fit and complexity, aiming for the lowest AIC value.

The AIC is given by:

\begin{equation}
\text{AIC} = 2k - 2\log(L),
\end{equation}

where $L$ is the likelihood of the model and $k$ is the number of parameters in the model, including the error variance $\sigma^2$. For a linear regression model with independent and identically distributed (iid) errors, $N(0, \sigma^2)$, fitted to $n$ observations, the log-likelihood can be written as:

\begin{equation}
\scalebox{0.85}{$
\log(L) = -\frac{n}{2} \log(2\pi) - \frac{n}{2} \log(\sigma^2) - \frac{1}{2\sigma^2} \sum_{i=1}^{n} \hat{e}_i^2, 
$}
\end{equation}

where $\hat{e}_i$ is the residual for the $i$th observation, and $\sigma^2$ is the variance of the errors. The AIC, in this context, becomes:

\begin{equation}
\scalebox{0.85}{$
\text{AIC} = 2k + n \log(2\pi) + n \log(\sigma^2) + \frac{1}{\sigma^2} \sum_{i=1}^{n} \hat{e}_i^2.
$}
\end{equation}

This form of the AIC balances the goodness of fit (as reflected by the residual sum of squares) and model complexity (as represented by $k$).

The operation of Bidirectional stepwise regression starts with either no predictors (forward selection) or all predictors (backward elimination), where the model iteratively adds or removes variables. Each step evaluates the impact on the AIC score. In forward selection, variables are added one by one, starting from the null model, such that the addition of each variable results in the largest decrease in AIC. In backward elimination, all variables are included in the model initially, and variables are removed one at a time, with the variable whose removal causes the smallest increase in AIC being dropped.

At each iteration, the change in AIC is computed as
$\Delta \text{AIC} = \text{AIC}_{\text{new}} - \text{AIC}_{\text{current}}$,
where $\text{AIC}_{\text{new}}$ refers to the AIC after adding or removing a variable, and $\text{AIC}_{\text{current}}$ is the AIC of the current model. If $\Delta \text{AIC} < 0$, the model is improved by the addition or removal of the variable. The process terminates when neither adding nor removing variables results in a lower AIC, signifying that the most parsimonious model, based on AIC, has been reached.

\subsection{Test results}
We operated bidirectional stepwise regression on both benchmarks individually and together to identify the factors that are significantly contributing to position bias. Specifically, the variables include lengths (input, output, and prompt), answer quality gap, LLM judges, candidate models, and task categories to predict position consistency and preference fairness respectively. 
Table \ref{tab:MTBench_PC_OLS}, \ref{tab:MTBench_PF_OLS} records the results of final step in stepwise regression for predicting $PC$ and $PF$, respectively. Table \ref{tab:DevBench_PC_OLS}, \ref{tab:DevBench_PF_OLS} serves for DevBench, and 
Table \ref{tab:both_benchmarks_PC_OLS}, \ref{tab:both_benchmarks_PF_OLS} is conducted on the integrated set of both benchmarks. The impact of variables on the model is ranked from highest to lowest, from bottom to top. Removed variables listed as None indicate the full model at this given step. 

Through benchmark testing, we verified that LLM judges, task categories, and the answer quality gap significantly contribute to position bias in terms of both position consistency and preference fairness. These findings align with our empirical results, showing that position bias varies notably by judge and task, with the answer quality gap being a key influencing factor. The extent of this impact is reflected by the magnitude of change in AIC when the given variable is removed. It is worth noting that while task output length remains a significant predictor for $PF$ and $PC$ in both benchmarks, the change in AIC magnitude after removing this variable is very minimal. This is consistent across both benchmarks individually and combined. We therefore conclude that, although position bias is influenced by task output length, this dependency is minimal.

\begin{table}[ht!]
    \centering
    \scalebox{0.8}{
    \begin{tabular}{r r r c c}
        \toprule
        Removed Variables & DF & Sum of Sq & RSS & AIC \\
        \midrule 
        None&  & & 163.75 & -18370 \\
        Task & 20 & 2.832 & 166.59 & -18319 \\
        Candidate & 38 & 4.472 & 168.23 & -18303 \\
        Quality gap & 1 & 21.953 & 185.71 & -17703 \\
        Judge & 13 & 55.417 & 219.17 & -16846 \\
        \bottomrule
    \end{tabular}
    }
    \caption{Final results of stepwise model selection for both benchmarks: Position Consistency}
    \label{tab:both_benchmarks_PC_OLS}
\end{table}

\begin{table}[ht!]
    \centering
    \scalebox{0.8}{
    \begin{tabular}{r r r c c}
        \toprule
        Removed Variables & DF & Sum of Sq & RSS & AIC \\
        \midrule
        None& & & 254.28 & -16103 \\
        Task output length & 1 & 0.836 & 255.12 & -16088 \\
        Quality gap & 1 & 11.339 & 265.62 & -15873 \\
        Task & 21 & 16.177 & 270.46 & -15817 \\
        Judge & 13 & 82.069 & 336.35 & -14641 \\
        \bottomrule
    \end{tabular}
    }
    \caption{Final results of stepwise model selection for both benchmarks: Preference Fairness}
    \label{tab:both_benchmarks_PF_OLS}
\end{table}

\begin{table}[ht!]
    \centering
    \scalebox{0.8}{
    \begin{tabular}{r r r c c}
        \toprule
        Removed Variables & DF & Sum of Sq & RSS & AIC \\
        \midrule
        None& & & 61.974 & -13312 \\
        Task output length & 1 & 0.0553 & 62.029 & -13311 \\
        Candidate & 29 & 1.6474 & 63.621 & -13282 \\
        Task & 7 & 1.5304 & 63.504 & -13244 \\
        Judge & 13 & 15.3637 & 77.338 & -12594 \\
        Quality gap & 1 & 15.6206 & 77.594 & -12559 \\
        \bottomrule
    \end{tabular}
    }
    \caption{Final results of stepwise model selection for MTBench: Position Consistency}
    \label{tab:MTBench_PC_OLS}
\end{table}

\begin{table}[ht!]
    \centering
    \scalebox{0.8}{
    \begin{tabular}{r r r c c}
        \toprule
        Removed Variables & DF & Sum of Sq & RSS & AIC \\
        \midrule
        None& & & 129.00 & -10909.2 \\
        Quality gap & 1 & 1.931 & 130.93 & -10861.3 \\
        Task & 7 & 9.295 & 138.29 & -10689.4 \\
        Judge & 13 & 58.847 & 187.85 & -9672.5 \\
        \bottomrule
    \end{tabular}
    }
    \caption{Final results of stepwise model selection for MTBench: Preference Fairness}
    \label{tab:MTBench_PF_OLS}
\end{table}

\begin{table}[ht!]
    \centering
    \scalebox{0.8}{
    \begin{tabular}{r r r c c}
        \toprule
        Removed Variables & DF & Sum of Sq & RSS & AIC \\
        \midrule
        None& & & 55.382 & -6940.2 \\
        Task output length & 1 & 0.257 & 55.638 & -6933.2 \\
        Candidate & 9 & 1.514 & 56.896 & -6905.4 \\
        Quality gap & 1 & 13.128 & 68.510 & -6525.3 \\
        Judge & 13 & 84.760 & 140.141 & -5146.6 \\
        \bottomrule
    \end{tabular}
    }
    \caption{Final results of stepwise model selection for DevBench: Position Consistency}
    \label{tab:DevBench_PC_OLS}
\end{table}

\begin{table}[ht!]
    \centering
    \scalebox{0.8}{
    \begin{tabular}{r r r c c}
        \toprule
        Removed Variables & DF & Sum of Sq & RSS & AIC \\
        \midrule
        None& & & 60.104 & -6753.9 \\
        Task output length & 1 & 0.061 & 60.165 & -6753.9 \\
        Candidate & 9 & 0.731 & 60.834 & -6748.2 \\
        Task & 13 & 1.305 & 61.408 & -6737.8 \\
        Quality gap & 1 & 1.783 & 61.886 & -6698.6 \\
        Judge & 13 & 80.875 & 140.979 & -5108.9 \\
        \bottomrule
    \end{tabular}
    }
    \caption{Final results of stepwise model selection for DevBench: Preference Fairness}
    \label{tab:DevBench_PF_OLS}
\end{table}

\section{Experiment Settings}
\label{appendix:experiment_settings}
This session specifies more detailed information about the judges, answer-generating models, tasks, and prompt templates used in this study. We choose to evaluate MTBench and DevBench for the following reasons: (1) all necessary information about the benchmark models, tasks, and questions is publicly available, making modifications convenient (2) they include a wide variety of answer-generating models, tasks, and task questions for a comprehensive evaluation (3) their human evaluations validated the reliability of state-of-the-art judging models (GPT-4 and GPT-4-Turbo) on their evaluation instances, hence model untested by prior work, if reaching high agreement with these validated judges, can be perceived reliable as well.

\subsection{Judges, Candidates, and Tasks}
\label{appendix:judges_models_tasks}

\paragraph{Judge} In this study, we choose seven \textbf{GPT}, four \textbf{Claude}, and three \textbf{Gemini} models as the judges. The specific versions for API call are specified as follows: o1-mini-2024-09-12 for \textbf{o1-mini}, gpt-4o-2024-05-13 for \textbf{GPT-4o}, gpt-4-1106-preview for \textbf{GPT-4-Turbo}, gpt-4-0613 for \textbf{GPT-4}, and gpt-3.5-turbo-1106 for \textbf{GPT-3.5-turbo}; claude-3-5-sonnet-20240620, claude-3-opus-20240229, claude-3-sonnet-20240229, and claude-3-haiku-20240307 for \textbf{Claude} series. The other model names and versions are as they are.

\paragraph{Model}
The reference (or baseline) answer-generating models are \textbf{vicuna-13b-v1.3} for MTBench and \textbf{human} for DevBench. They are chosen to ensure a baseline quality of responses and an expected widely spread quality gap across evaluations. The other models that are compared to the reference models, namely ``Model" in our context, are listed as follows.

\begin{itemize}[leftmargin=*]
    \item \textbf{MTBench (30)}: alpaca-13b, baize-v2-13b, chatglm-6b, claude-instant-v1, claude-v1, dolly-v2-12b, falcon-40b-instruct, fastchat-t5-3b, gpt-3.5-turbo, gpt-4, gpt4all-13b-snoozy, guanaco-33b, guanaco-65b, h2ogpt-oasst-open-llama-13b, koala-13b, llama-13b, mpt-30b-chat, mpt-30b-instruct, mpt-7b-chat, nous-hermes-13b, oasst-sft-4-pythia-12b, oasst-sft-7-llama-30b, palm-2-chat-bison-001, rwkv-4-raven-14b, stablelm-tuned-alpha-7b, tulu-30b, vicuna-33b-v1.3, vicuna-7b-v1.3, wizardlm-13b, wizardlm-30b
    \item \textbf{DevBench (10)}: codellama-7b-instruct, codellama-13b-instruct, codellama-34b-instruct, deepseek-coder-1.3b-instruct, deepseek-coder-6.7b-instruct, deepseek-coder-33b-instruct, gpt-3.5-turbo-1106, gpt-4-0125-preview, gpt-4-0613, gpt-4-1106-preview
\end{itemize}

The model names are exactly what MTBench \cite{zheng2024judging} and DevBench \cite{li2024devbench} used in their studies. That is why for GPTs, DevBench specifies the exact version (e.g., gpt-4-0613) while MTBench doesn't (e.g., gpt-4). In this study, we directly use the provided answers of these models to the task questions to form answer pairs and queries for the LLM judges.

\paragraph{Task}
For tasks, we also follow the original studies of these two benchmarks, except for DevBench we separate the \textit{gerenal}
metrics into detailed ones and considered them as different tasks. In this sense, our study experiments on the following tasks to provide a comprehensive study on the positon bias of LLM-as-a-Judge:

\begin{itemize}[leftmargin=*]
    \item \textbf{MTBench (8)}: coding, extraction, humanities, math, reasoning, roleplay, stem, and writing.
    \item \textbf{Devbench (14)}: 
    \begin{itemize}
        \item \textbf{UML class} (4): cohesion\_and\_decoupling, complexity, practicability, and faithfulness
        \item \textbf{UML sequence} (5): cohesion\_and\_decoupling, interaction\_complexity, practicability, uniformity\_and\_integration, and faithfulness
        \item \textbf{architecture design} (5): conformance, design\_and\_coding, practicability, uniformity\_and\_integration, and faithfulness
    \end{itemize}
\end{itemize}

\subsection{Prompt Settings}
\label{appendix:prompt_settings}
We follow the original prompt settings of MTBench and DevBench in our study of pairwise comparative LLM-as-a-Judge.

Though written differently, these prompts all share same key components: 
\begin{itemize}[leftmargin=*]
    \item A system prompt explaining the judging task and the role the LLM should be playing.
    \item Emphasized ``should" and``shouldn't"s.
    \item A prompt structure with placeholders for specific questions and model answers
    \item A specified output format for later judgment extraction
    \item Chain-of-Thought \cite{wei2022chain} prompts requiring the LLM judge to provide reasons for its judgment
\end{itemize}

The detailed prompt templates are specified below.

\begin{figure*}[ht]
    \centering
    \includegraphics[width=\textwidth]{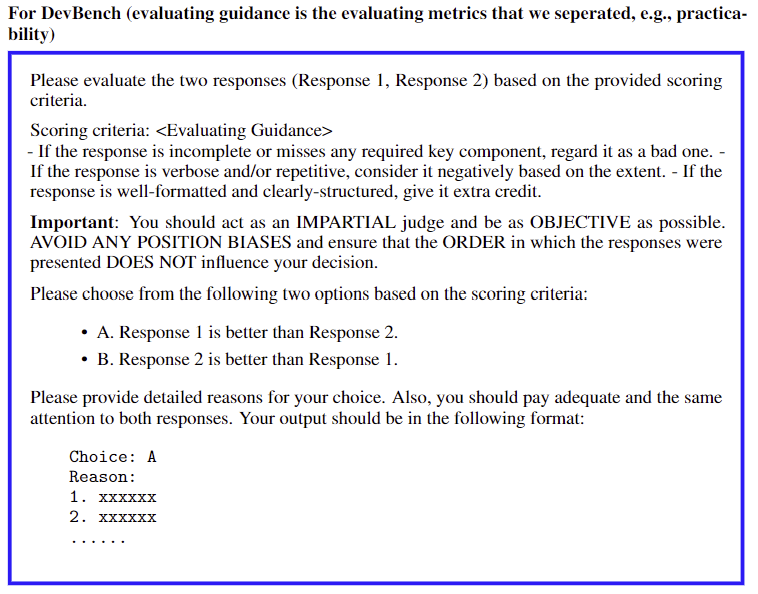}
\end{figure*}

\clearpage

\begin{figure*}[ht]
    \centering
    \scalebox{1.0}{
    \includegraphics[width=\textwidth]{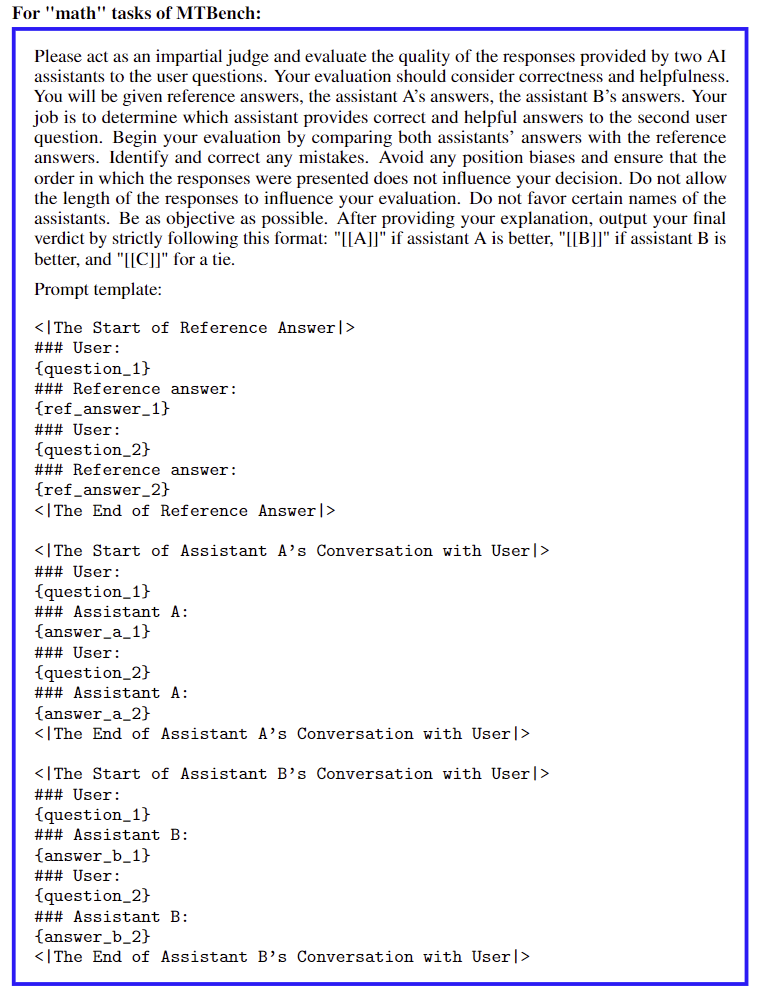}
    }
\end{figure*}

\clearpage
\begin{figure*}[ht]
    \centering
    \includegraphics[width=\textwidth]{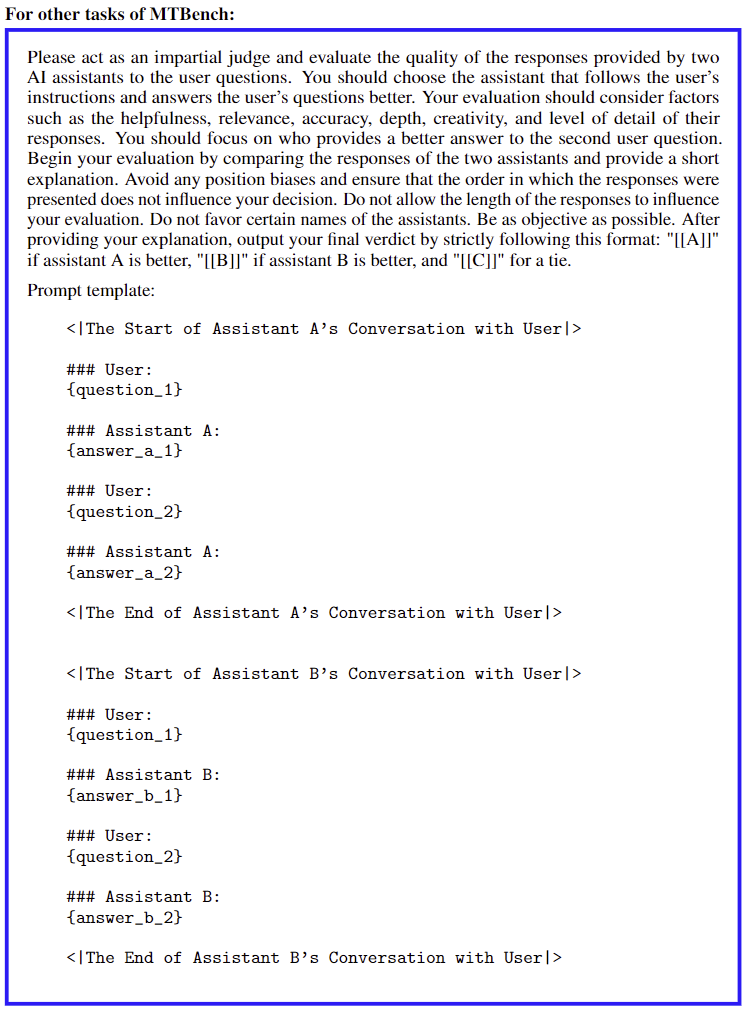}
\end{figure*}

\clearpage

    
    
    
    

\end{document}